\documentclass[final]{cvpr}

\usepackage{times}
\usepackage{epsfig}
\usepackage{graphicx}
\usepackage{amsmath}
\usepackage{amsfonts}
\usepackage{nicefrac}       
\usepackage{microtype}      
\usepackage{amssymb}

\usepackage[ruled,vlined]{algorithm2e}

\SetCommentSty{mycommfont}

\usepackage{graphicx}
\usepackage{caption}
\usepackage{subcaption}
\usepackage{mathtools}
\usepackage{floatrow}
\usepackage{color}
\usepackage{setspace}
\usepackage{enumitem}
\usepackage{ctable}
\usepackage{soul}
\usepackage{multirow}
\usepackage[toc,page]{appendix}

\usepackage{ifthen}

\usepackage[pagebackref=true,breaklinks=true,colorlinks,bookmarks=false]{hyperref}

\newcommand{\FpnasEffnetSpeedup}[0]{132}
\newcommand{\Lzerosearchtime}[0]{28.7 }

\def\cvprPaperID{2777} 
\def\confYear{CVPR 2021}

\begin{document}

\title{FP-NAS: Fast Probabilistic Neural Architecture Search}

\author{
    Zhicheng Yan\thanks{Correspondence to Zhicheng Yan $<$zyan3@fb.com$>$.}, Xiaoliang Dai, Peizhao Zhang, Yuandong Tian, Bichen Wu, Matt Feiszli\\
    Facebook AI\\
}

\maketitle
\pagestyle{empty}
\thispagestyle{empty}

\begin{abstract}

Differential Neural Architecture Search (NAS) requires all layer choices to be held in memory simultaneously; this limits the size of both search space and final architecture.  In contrast, Probabilistic NAS, such as PARSEC, learns a distribution over high-performing architectures, and uses only as much memory as needed to train a single model. Nevertheless, it needs to sample many architectures, making it computationally expensive for searching in an extensive space. To solve these problems, we propose a sampling method adaptive to the distribution entropy, drawing more samples to encourage explorations at the beginning, and reducing samples as learning proceeds. Furthermore, to search fast in the multi-variate space, we propose a coarse-to-fine strategy by using a factorized distribution at the beginning which can reduce the number of architecture parameters by over an order of magnitude. We call this method \textit{Fast Probabilistic NAS (FP-NAS)}. Compared with PARSEC, it can sample $64 \%$ fewer architectures and search \textbf{2.1}$\times$ faster. Compared with FBNetV2, FP-NAS is \textbf{1.9}$\times$ - \textbf{3.5}$\times$ faster, and the searched models outperform FBNetV2 models on ImageNet. FP-NAS allows us to expand the giant FBNetV2 space to be wider (i.e. larger channel choices) and deeper (i.e. more blocks), while adding Split-Attention block and enabling the search over the number of splits. When searching a model of size $0.4G$ FLOPS, FP-NAS is \textbf{132}$\times$ faster than EfficientNet, and the searched FP-NAS-L0 model outperforms EfficientNet-B0 by $0.7\%$ accuracy. Without using any architecture surrogate or scaling tricks, we directly search large models up to $1.0G$ FLOPS. Our \textit{FP-NAS-L2} model with simple distillation outperforms BigNAS-XL with advanced inplace distillation by $0.7\%$ accuracy using similar FLOPS.

\end{abstract}

\section{Introduction}


Designing efficient architectures for visual recognition requires extensive exploration in the search space; doing this by hand takes substantial human effort and computational resources. Since the introduction of AlexNet~\cite{2012alexnet}, handed-crafted models like ResNet~\cite{2016resnet}, Densenet~\cite{2017densenet} and InceptionV4~\cite{2017inceptionv4}, have become increasingly deep with more complex connectivity.  Exploring this space manually is difficult and time-consuming.  Neural Architecture Search (NAS) aims to automate the architecture design process. However, early approaches based on evolution or reinforcement learning take hundreds or thousands of GPU days just for CIFAR10 or Imagenet target datasets~\cite{2016nas, 2019nasnet, 2019Mnasnet, 2019AmoebaNet}. 

Recently, differentiable neural architecture search (DNAS)~\cite{2018darts} relaxed the discrete representation of architectures to a continuous space, enabling efficient search with gradient descent. This comes at a price: DNAS instantiates all layer choices in memory, and computes all feature maps. 
Therefore, its memory footprint increases linearly with the number of layer choices, and greatly limits the size of both search space and final architecture. PARSEC~\cite{2019pnas} presents a probabilistic version of DNAS which samples individual architectures from a distribution on search space.  PARSEC's sampling strategy uses much less memory, but it samples a large number of architectures to fully explore the space, which significantly increases computational cost.  Here we pose the following question: \textit{Can we reduce PARSEC's compute cost and maintain a small memory footprint to support the fast search of both small and large models?}

In this work, we accelerate PARSEC by proposing two novel techniques. First, we replace the fixed architecture sampling with a dynamic sampling adaptive to the entropy of architecture distribution. In particular, we sample more architectures in the early stage to encourage exploration, and fewer later, as the distribution concentrates on a smaller set of promising architectures. Furthermore, in multi-variate space where several variables (e.g. block type, feature channel, expansion ratio) are searched over, we propose a coarse-to-fine search strategy. In the beginning stage, we adopt a factorized distribution representation to search at a coarse-grained granularity, which uses much fewer architecture parameters and makes the learning much faster. In the following stage, we seamlessly convert the factorized distribution into the joint distribution to support fine-grained search.

\begin{figure}[t!]
\vspace{-0.9cm}
\centering
\subfloat{
\includegraphics[width=0.86\columnwidth]{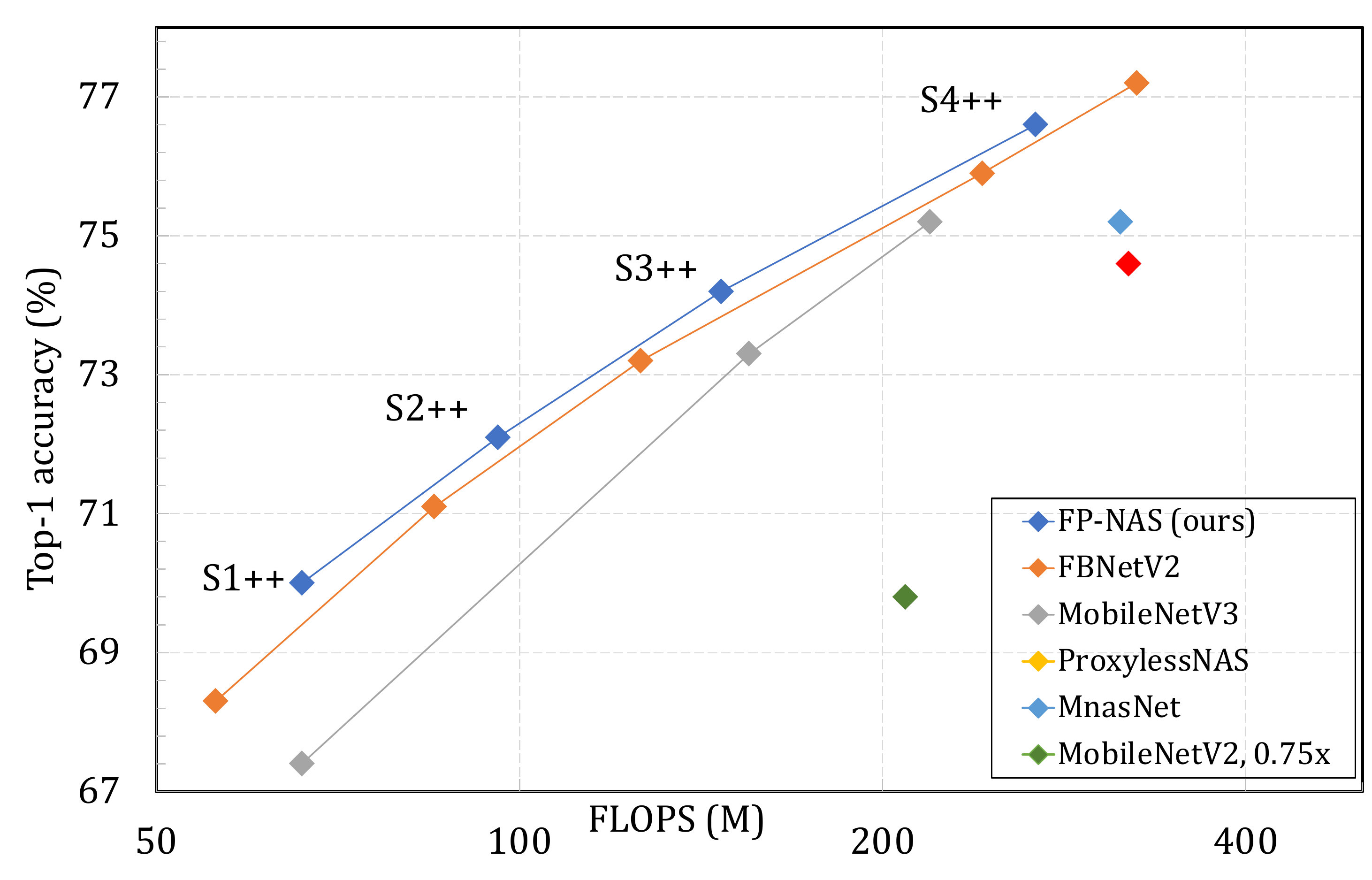}
}
\\
\subfloat{
\includegraphics[width=0.87\columnwidth]{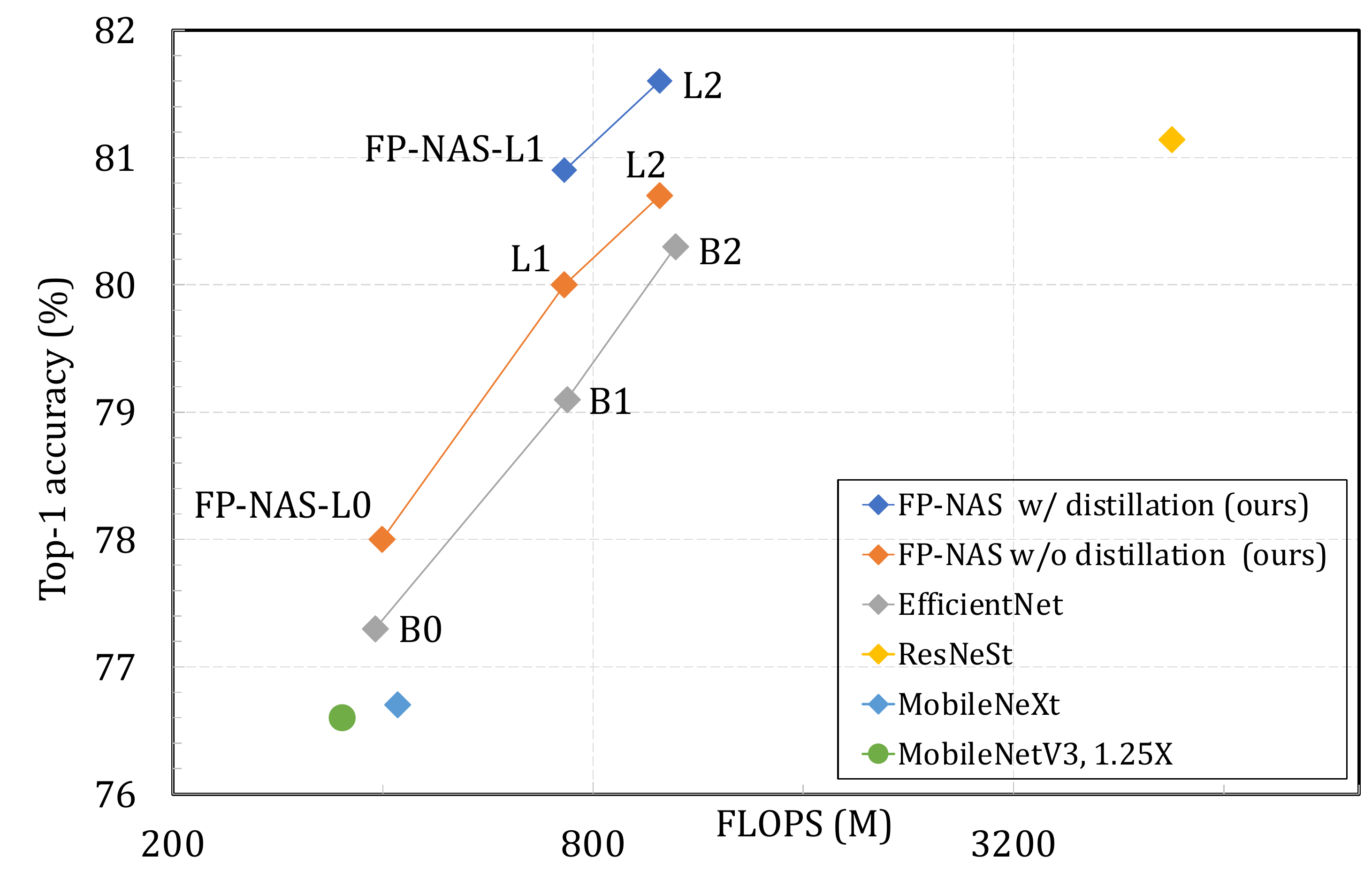}
}
\caption{\textbf{Model FLOPS vs. ImageNet validation accuracy}. All numbers are obtained using a single-crop and a single model. \textbf{Top}: Small models using less than 350M FLOPS. 
\textbf{Bottom}: Large models. Our searched models, including both small- (i.e. FP-NAS-S++) and large models (i.e. FP-NAS-L), achieve better accuracy-to-complexity trade-off than other models. For example, when model is trained from scratch, our FP-NAS-L0 model achieves $0.7\%$ higher accuracy than EfficientNet-B0 while searching it is over $\FpnasEffnetSpeedup\times$ faster (\Lzerosearchtime  Vs. 3790+ GPU-days). We also use notation \textit{(+ distillation)} to report the results of our large models with vanilla model distillation~\cite{hinton2015distilling}.
}
\label{fig:flops_acc_plot}
\vspace{-0.3cm}
\end{figure}

When searching in the FBNetV2 space~\cite{2019FBNetV2}, FP-NAS samples $64\%$ fewer architectures and searches $2.1\times$ faster compared with the PARSEC method. Compared with FBNetV2, FP-NAS is $3.5\times$ faster, and the searched models outperform FBNetV2 models on ImageNet. To further demonstrate FP-NAS' efficiency, we expand the large FBNetV2 search space, and introduce searchable Split-Attention blocks~\cite{2020resnest} which increases the space size by over $10^3\times$. Our main results are shown in Fig~\ref{fig:flops_acc_plot}. In total, we searched a family of FP-NAS models, including 3 large architectures (\textit{from L0 to L2}) of over $350M$ FLOPS and 4 small ones (\textit{from S1++ to S4++}). When searching models using 0.4G target FLOPS, FP-NAS only uses 28.7 GPU-days, and \textit{is at least \FpnasEffnetSpeedup $\times$ faster than the search of EfficientNet-B0}, which uses a similar amount of FLOPS. Moreover, our method also discovers more superior FP-NAS-L0 model which outperforms EfficientNet-B0 by $0.7\%$ top-1 accuracy. The largest model FP-NAS-L2, found via direct search on a subset of ImageNet, uses $1.05G$ FLOPS for a single crop. To our knowledge, this is the largest architecture obtained via direct search.

To summarize, this work makes the contributions below:
\begin{itemize}[leftmargin=*]

\item An adaptive sampling method for fast probabilistic NAS, which can sample $60\%$ fewer architectures and accelerate the search in FBNetV2 space by $1.8\times$.

\item A coarse-to-fine search method by adopting a factorized distribution representation with much fewer architecture parameters in the early coarse-grained search stage. It can further accelerate the overall search by $1.2\times$.

\item For searching small models, comparing with FBNetV2, FP-NAS is not only $3.5\times$ faster, but also discovers models with substantially better accuracy-to-complexity trade-off.

\item For searching large models, compared with EfficientNet, when searching models of $0.4G$ FLOPS, FP-NAS is not only $\FpnasEffnetSpeedup\times$ faster, but also discovers a model with $0.7\%$ higher accuracy on ImageNet. The largest model we searched is FP-NAS-L2, which uses $1.05G$ FLOPS, outperforms EfficientNet-B2 by $0.4\%$ top-1 accuracy, while the search cost is much lower.

\item We expand FBNetV2 search space by replacing Squeeze-Excite module~\cite{2018senet} with searchable Split-Attention (SA) module~\cite{2020resnest}. We demonstrate uniformly inserting SA modules to the model with a fixed number of splits, as done in hand-crafted ResNeSt~\cite{2020resnest} models, is sub-optimal, and models with searched SA modules are more competitive.

\end{itemize}

\section{Related Work}

\noindent{\textbf{Hand-Crafted Models.}} The conventional way of building ConvNet is to design repeatable building blocks, and stack them to form deep models, including ResNet~\cite{2016resnet, 2017resnext}, DenseNet~\cite{2017densenet}, and Inception~\cite{2016label-smooth, 2015inceptionv1, 2017inceptionv4}. Meanwhile, manually designing compact mobile models has also attracted lots of interest due the prevalence of mobile devices. Mobile models uses more compute-efficient blocks, such as inverted residual block~\cite{2018mobilenetv2} and shuffling layers~\cite{2018shufflenet, 2018shufflenetv2}. However, recent models discovered by neural architecture search have surpassed hand-crafted models in various tasks~\cite{du2020spinenet, 2019efficientnet, bender2018understanding, 2018param-share-nas}.

\noindent{\textbf{Non-Differentiable Neural Architecture Search.}} Early NAS methods are based on either reinforcement learning (RL)~\cite{2019Mnasnet} or evolution~\cite{2017large-scale-evolution}. In the pioneering work~\cite{2016nas}, a RNN controller is adopted to sample architectures, which are trained to obtain accuracy as the reward signal for updating the controller. It requires to train tens of thousands of architectures, which is computationally prohibitive. Similarly, in NASNet~\cite{2019nasnet}, it takes 2,000 GPU days to search architectures for CIFAR10 and ImageNet. In the AmoebaNet~\cite{2019AmoebaNet}, which is based on evolution, the search algorithm iteratively evaluates a small number of child architectures evolved from the best-performing architectures in the population to speed up the search, but still requires to train thousands of individual architectures. Recently, EfficientNet~\cite{2019efficientnet} built big models by scaling up the small ones from RL-based search~\cite{2019Mnasnet} jointly along the depth, width and input resolution. BigNAS~\cite{bignas} trains a single-stage model with inplace distillation, and induce child models of different sizes without retraining or fine-tuning. In this work, with the proposed fast probabilistic NAS method, we show directly searched big architectures without any scaling trick can achieve a better accuracy-to-complexity trade-off.

\noindent{\textbf{Differentiable Neural Architecture Search.}} DARTS~\cite{2018darts} relaxes the discrete search space to be continuous, and optimizes the architecture by gradient descent. While being much faster, it requires to instantiate all layer choices in the memory, making it difficult to directly search big architectures in large space. Therefore, DARTS needs to use a shallow version of model at search time to serve as the surrogate, and repeats the searched cells many more times at evaluation time to build larger models.  

Following works improve DARTS by path pruning to reduce memory footprint as in ProxylessNAS~\cite{2018proxylessnas}, more fine-grained search space~\cite{2019atomnas}, hierarchical search space~\cite{2017hierarchical-nas}, better optimizer~\cite{2019xnas}, better architecture sampler~\cite{2019DATA-NAS}, being platform-aware~\cite{2018FBNet, 2019chamnet}, and searching over channels and input resolution in a memory efficient manner~\cite{2019FBNetV2}. In GDAS~\cite{2019oneshot4hr} paper, a differentiable sampler based on Gumbel-Max trick~\cite{2016gumbel-trick} is proposed to only sample one architecture at a time. This reduces the memory usage but the searched architectures have performance inferior to those searched by evolution-based methods~\cite{2019AmoebaNet}. PARSEC~\cite{2019pnas} proposes a sampling-based method to learn a probability distribution over architectures, and is also memory-efficient. However, to achieve good search results, it needs to constantly sample a large number of architectures, which is computationally expensive. In this work, we propose to adaptively reduce the architecture samples based on entropy of architecture distribution, substantially reduce the search time, and enable the search of bigger architectures.

For searching mobile models, differentiable NAS methods are adapted to be hardware-aware, considering model cost, such as FLOPS, memory, latency on specific hardware~\cite{2018FBNet, 2019FBNetV2, 2019chamnet, 2019Mnasnet, 2019mobilenetv3}. In this work, we adopt a hinge-linear penalty on the model FLOPS to constrain the computational cost and support the search of models with target FLOPS. 

\section{Fast Probabilistic NAS}

\subsection{Background}

Our method extends PARSEC~\cite{2019pnas} (a probabilistic NAS), which we briefly review here. In DNAS~\cite{2018darts}, for each layer $l$ we have a set of candidate operations $\mathit{O}$; each operation $\mathit{o}(\cdot)$ can be applied to input feature $x^l$.  Discrete choice is relaxed to a weighted sum of candidate operations:

\vspace{-0.2cm}
\begin{equation}
\mathbf{x}^{l+1} = \sum_{o \in O } \frac{exp(\alpha_o^l)}{\sum_{o' \in O} exp(\alpha_{o'}^l)}o(\mathbf{x}^l)
\end{equation}
where $\{ \alpha_o^l \}_{o \in \mathit{O} }$ denotes architecture parameters at layer $l$. 

An architecture $\mathbf{A}$ is uniquely defined by the individual choices at $L$ layers $\mathbf{A}=(A^1,...,A^L)$. In PARSEC, a prior distribution $p(\mathbf{A} | \mathbf{\alpha})$ on the choices of layer operation is introduced, where architecture parameters $\mathbf{\alpha}$ denote the probabilities of choosing different operations. Individual architectures can be represented as discrete choices of $\{A^l\}$ and sampled from $p(\mathbf{A} | \mathbf{\alpha})$. Therefore, architecture search is transformed into learning the distribution $p(\mathbf{A} | \mathbf{\alpha})$ under certain supervision. For simplicity, we first assume the choices at different layers are independent to each other, and the probability of sampling an architecture $\mathbf{A}$ is shown below. 

\begin{equation}
\begin{aligned}
 P(\mathbf{A}|\mathbf{\alpha}) = \prod_l P(A^l | \mathbf{\alpha}^l)
\end{aligned}
\end{equation}

For image classification where we have images $\mathbf{X}$ and labels $\mathbf{y}$, probabilistic NAS can be formulated as optimizing the continuous  architecture parameters $\alpha$ via an empirical Bayes Monte Carlo procedure~\cite{2000monte-carlo} 

\vspace{-0.2cm}
\begin{equation}
\begin{aligned}
P(\mathbf{y} | \mathbf{X},   \mathbf{\omega}, \alpha ) = \int P(\mathbf{y} | \mathbf{X}, \mathbf{\omega}, \mathbf{A}) P(\mathbf{A}|\mathbf{\alpha}) \textup{d}\mathbf{A} \\
\approx \frac{1}{K} \sum_{k} P(\mathbf{y} | \mathbf{X}, \mathbf{\omega}, \mathbf{A}_{k})
\end{aligned}
\end{equation}
where $\omega$ denotes the model weights. The continuous integral of data likelihood is approximated by sampling architectures and averaging the data likelihoods from them. We can jointly optimize architecture parameters $\mathbf{\alpha}$ and model weights $\mathbf{\omega}$ by estimating gradients $\nabla _{\alpha}\;\textup{log}\;P(\mathbf{y}|\mathbf{X}, \omega, \mathbf{\alpha} )$ and $\nabla _{w} \;\textup{log}\;P(\mathbf{y}|\mathbf{X}, \omega, \mathbf{\alpha})$  through the sampled architectures.

To reduce over-fitting, separate training- and validation set are used to compute gradients w.r.t $\mathbf{\alpha} $ and $\mathbf{\omega}$. The probabilistic NAS proceeds in an iterative way. In each iteration, $K$ architecture samples $\{\mathbf{A}_k\}_{k=1}^{K}$  are drawn from $P(\mathbf{A} | \mathbf{\alpha})$. For sampled architectures, gradients w.r.t $\mathbf{\alpha} $ and $\mathbf{\omega}$ in sampled operations are computed on a batch of training data.

\subsection{Adaptive Architecture Sampling}
In PARSEC~\cite{2019pnas}, a fixed number of architectures are sampled during the entire search to estimate the gradients. For example, to search cell structures in the DARTS~\cite{2018darts} space on CIFAR10 (a relatively small search space), the number of samples is held fixed at 16. Such choice is ad-hoc and could be suboptimal for searching in spaces of different size. In the beginning of the search where the architecture distribution 
$P(\mathbf{A} | \mathbf{\alpha})$ is flat, a larger number of samples are needed to approximate the gradients. As the search proceeds, the distribution concentrates mass on a small set of candidates. In such case, we can reduce the search computation by drawing fewer samples. \\

Formally, we propose a simple yet effective sampling method adaptive to the learning of architecture distribution. During the search, we adjust the size of architecture samples $K$ to be proportional to the entropy of $P(\mathbf{A} | \mathbf{\alpha} )$.  Early in the search, entropy is high, encouraging more exploration.  Later, entropy decreases as a subset of candidate operations are deemed to be more promising, and the sampling can be more biased towards them.  Specifically, we set
\begin{equation}
    K = \lambda\; H(P(\mathbf{A} | \mathbf{\alpha}))
\end{equation}
where $H$ denotes the distribution entropy, and $\lambda$ a predefined scaling factor. In Section~\ref{sec:adaptive_sampling}, we show adaptive sampling can greatly reduce the search time without degrading the searched model. The choice of $\lambda$ is  discussed in Section~\ref{sec:on_arch_sample_size}.

\subsection{Coarse-to-Fine Search in Multi-Variate Space}
The search space of each layer operation $A^l$ can include multiple search variables, such as convolution kernel size, nonlinearity and feature channel. In such multi-variate space, when we use a vanilla joint distribution (JD) representation, the number of architecture parameters is a product of cardinalities of individual variables, which grows rapidly as more variables are added. For example, the search space used later in this work (See Table~\ref{tab:macro-structure} and ~\ref{tab:micro-structure}) has 5 variables, including kernel size, nonlinearity, Squeeze-Excite, expansion rate in MobileNetV3~\cite{2019mobilenetv3} and channel. When their individual cardinalities are 3, 2, 2, 6, and 10 respectively, the JD uses $ prod([3, 2, 2, 6, 10])= 720$ parameters. We can factorize the large JD, and obtain a more compact representation using multiple small distributions. For the 5-dimensional search space above, we use 5 small distributions, and the total architecture parameters can be dramatically reduced to $sum([3, 2, 2, 6, 10])= 23$, which is over $31\times$ less.

Formally, in a search space of layer operation $A^l$ with $M$ search variables, each layer can be represented as a \textit{M}-tuple $A^l = (A^l_{1}, ..., A^l_{M}) $. We adopt a factorized distribution (FD) for the layer operation below.

\begin{equation}
\begin{aligned}
P(A^l|\mathbf{\alpha}^l) = P(\{A^l_m\}_{m=1}^M | \{\alpha^l_m \}_{m=1}^M  ) \\
= \prod_{m=1}^M P(A^l_m | \alpha^l_m ) \;\;\;\; \textup{where} \  A^l_m \in D^l_m
\end{aligned}
\label{eqn:factorized_dist}
\end{equation}
Here, $D^l_m$ denotes the set of choices for variable $A^l_m$. Compared with JD, FD greatly reduces the total architecture parameters from $\prod_{m} \left | D^l_m \right |$ to $\sum_{m} \left | D^l_m \right |$, which often leads to more than an order of magnitude reduction in practice, and can greatly accelerate the search. However, FD ignores the correlation between search variables, and can only support coarse-grained search. For example, the search of expansion rate and channel is likely to be correlated since the inner channel within the MBConv block as in MobileNetV3~\cite{2019mobilenetv3} is a product of expansion rate and channel. A large expansion rate might be more preferred when the channel is not high, but can be less preferred when the channel is already high because it can introduce an excessive amount of FLOPS but does not improve the classification accuracy.

To support fast search, we propose a coarse-to-fine search method by using a schedule of mixed distributions which starts the search with FD for a number of epochs, and later converts FD to the JD for the following epochs. As shown in Section~\ref{sec:coarse_fine_search}, the coarse-to-fine search can accelerate the search without compromising the performance of the searched model.

\subsection{Architecture Cost Aware Search}

Without any constraint on the architecture cost (e.g. FLOPS, parameters or latency), the search tends to favor big architectures, which are more likely to fit training data better but might be not suitable for efficiency-sensitive applications.  To search architectures with a target cost in mind, we adopt a hinge loss, which penalizes architectures when they use more than the target cost. We use FLOPS as the model cost in this work, but other choices, such as latency, can also be used.  Our full cost-aware loss consists of the data likelihood and the model cost.

\begin{equation}
\begin{split}
    \mathcal{L}(\omega, \mathbf{\alpha}) &= -\textup{log}\;P(\mathbf{y} | \textbf{X}, \omega, \mathbf{\alpha}) + \beta\;\textup{log}\;C(\mathbf{\alpha}) \\
    C(\mathbf{\alpha}) &= \int C(\mathbf{A})P(\mathbf{A}|\mathbf{\alpha}) \textup{d}\alpha  \approx \frac{1}{K} \sum_{k=1}^{K} C(\mathbf{A}_k) 
\end{split}
\label{eqn:hinge-loss-cost}
\end{equation}
where the hinge cost for a sampled architecture is $C(\mathbf{A}_k) = max(0, \frac{\textup{FLOPS}(\mathbf{A}_k)}{\textup{TAR}} - 1)$, $\beta$ denotes the coefficient of architecture cost, and $C(\mathbf{\alpha})$ the expected  architecture cost, which can be estimated by averaging the costs of sampled architectures. The gradient w.r.t $\mathbf{\alpha}$ is shown below.

\begin{equation}
\nabla _{\mathbf{\alpha}}\;\mathcal{L}(\omega, \mathbf{\alpha}) \approx \sum_{k=1}^K m_k \nabla _{\mathbf{\alpha}}\;-\textup{log}\;P(\mathbf{A}_k | \mathbf{\alpha})
\label{eqn:grad-cost-aware-loss}
\end{equation}
where $m_k = \frac{P(\mathbf{y}_{val} | \mathbf{X}_{val},\;\omega,\;\mathbf{A}_k)}{\sum _{k}\;P(\mathbf{y}_{val} | \mathbf{X}_{val},\;\omega,\;\mathbf{A}_{k})} - \beta\frac{C(\mathbf{A}_k)}{\sum_{k} C(\mathbf{A}_{k})}$, and denotes cost-aware architecture important weights. Intuitively, architecture parameters $\mathbf{\alpha}$ are updated to bias towards those architectures which both achieve high data likelihood on the validation data and use low FLOPS. At the end of the search, we select the most probable one in the learned distribution.

\section{Search Spaces}

\begin{table}[t!]
\vspace{-0.3cm}
\centering
\resizebox{0.99\columnwidth}{!}{

  \begin{tabular}{cccccc}
   Max Input ($S^2\times C$) &  Operator   & Expansion &    Channel & Repeat & Stride      \\
    
    \specialrule{.15em}{.1em}{.1em}

  $224^2 \times 3$  &  conv $3\times3$ & 1 & 16  &   1  &   2  \\
  
  $112^2\times 16$   & MBConv   &   1  &   (12, 16, 4)  & 1 & 1   \\
  $112^2\times 16$   & MBConv   &   (0.75, 4.5, 0.75)  &   (16, 24, 4)  & 1 & 2   \\
  
  $56^2\times24$  & MBConv   &   (0.75, 4.5, 0.75)  &   (16, 24, 4)  & 2 & 1   \\
  $56^2\times24$   & MBConv   &   (0.75, 4.5, 0.75)  &   (16, 40, 8)  & 1 & 2  \\
  
  $28^2\times40$   & MBConv   &   (0.75, 4.5, 0.75)  &   (16, 40, 8)  & 2 & 1  \\
  $28^2\times40$   & MBConv  &    (0.75, 4.5, 0.75)  &   (48, 80, 8)  & 1 & 2  \\
  
  $14^2\times80$   & MBConv   &   (0.75, 4.5, 0.75)  &   (48, 80, 8)  & 2 & 1  \\
  $14^2\times80$  & MBConv     &    (0.75, 4.5, 0.75)  &   (72, 112, 8)  & 3 & 1 \\
  $14^2\times112$  & MBConv     &    (0.75, 4.5, 0.75)  &   (112, 184, 8)  & 1 & 2 \\
  
  $7^2\times184$   & MBConv  & (0.75, 4.5, 0.75) & (112, 184, 8) & 3 & 1 \\
  $7^2\times184$ & conv $1\times1$ & - & 1984 & 1 & 1 \\
  
  $7^2\times1984$ & avgpool & - & - & 1 & 1 \\
  1984 & fc & - & 1000 & 1 & -  \\
\specialrule{.1em}{.05em}{.05em}
  \end{tabular}
}
\caption{\textbf{FBNetV2-F macro architecture}. Each row represents a block group. \textit{MBConv} denotes the inverted residual block in MobileNetV2~\cite{2018mobilenetv2}. \textit{Expansion} and \textit{Channel} denote expansion rate and the output channel of the block. Their search range is denoted as \textit{(min, max, step)}. \textit{Repeat} denotes the repeating times of the block, and \textit{stride} means the stride of first one among them. 
}
\label{tab:macro-structure}
\end{table}

\begin{figure}[t!]
\vspace{-0.2cm}
   \centering
   \includegraphics[width=.9\columnwidth]{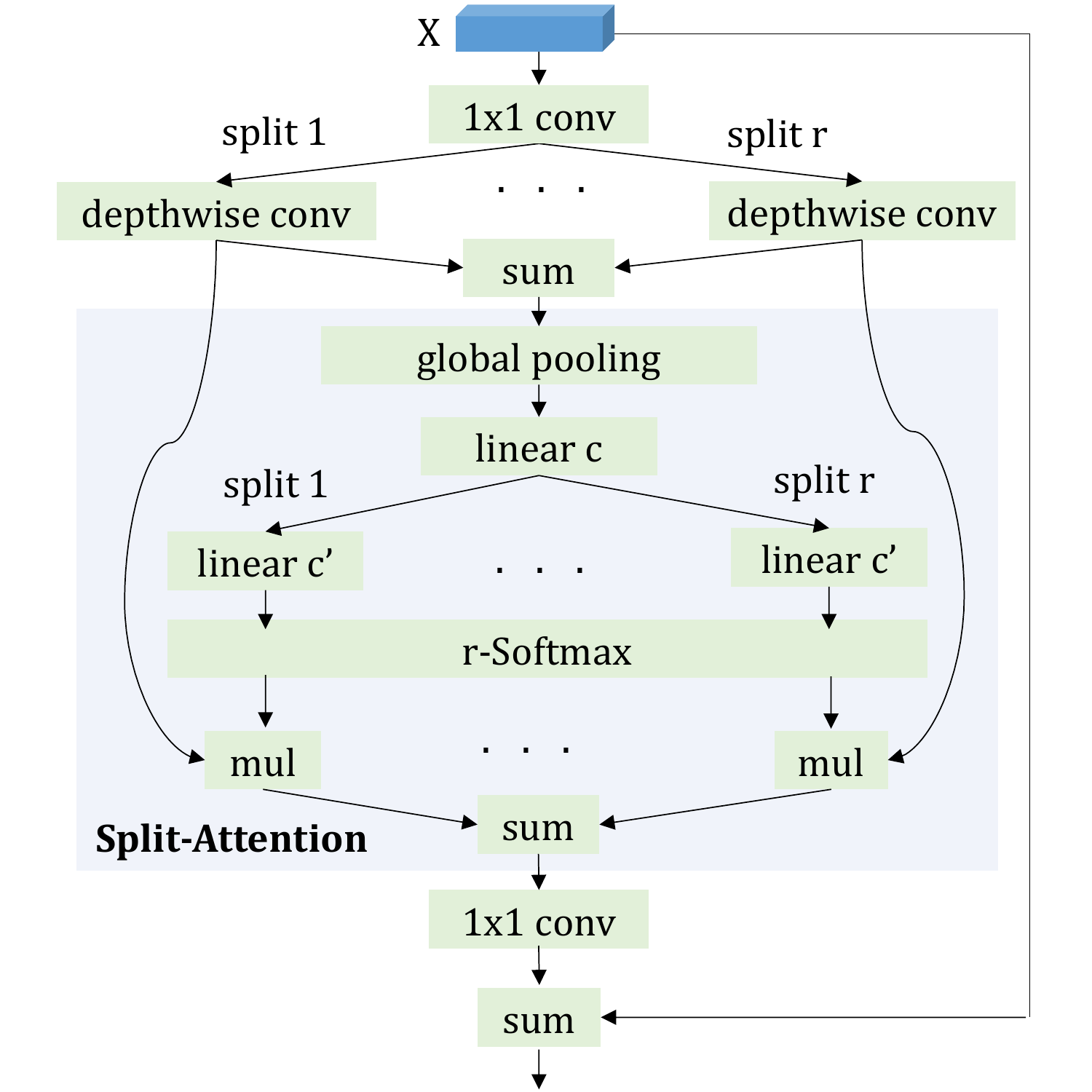}
   \caption{\textbf{MBConv block with searchable Split-Attention module.} 
   }
   \label{fig:MBConv_SA}
\vspace{-0.3cm}
\end{figure}

We consider 4 difference spaces below to search models.

\noindent{\textbf{FBNetV2-F space~\cite{2019FBNetV2}.}} We conduct most ablation studies in this space, which is defined by the macro-architecture in Table~\ref{tab:macro-structure},
and the micro-architecture in the 1st row of Table~\ref{tab:micro-structure}. It has multiple search variables, including convolution kernel size, nonlinearity type, the use of Squeeze-Excite block~\cite{2018senet}, block expansion rate, and block feature channel, and contains $6\times10^{25}$ different architectures.

\noindent{\textbf{FBNetV2-F-Fine space.}} The difference from FBNetV2-F space is each MBConv block is allowed to have different micro-architecture. FBNetV2-F-Fine contains $1\times10^{45}$ architectures, which is $10^{19}\times$ larger than FBNetV2-F, and can be viewed as a fine-grained version of FBNetV2-F space.

\noindent{\textbf{FBNetV2-F++ space.}} To demonstrate the search efficiency of our method, we extend the micro-architecture by replacing Squeeze-Excite (SE) module with Split-Attention (SA) module~\cite{2020resnest} in the MBConv block (Fig~\ref{fig:MBConv_SA}), and denote it as FP-NAS micro-architecture (Table~\ref{tab:micro-structure}, 2nd row). SA module generalizes SE module from one split to multiple splits. However, in the original hand-crafted ResNeSt models~\cite{2020resnest}, a fixed number of splits (e.g. 2 or 4) is chosen, and SA modules are used within all ResNeXt blocks. We hypothesize it is unnecessary to use SA module everywhere, which will incur computational overhead. Therefore, we make SA module fully searchable by extending the search variable \textit{no. of splits} to have extra choices $\{2, 4\}$, which means each block group can independently choose whether SA module is used and how many splits to use. Note we do not share the model weights of MBConv block between choices of \textit{no. of splits}, which means the total model weights of the supernet will double as extra choices $\{2, 4\}$ are introduced, and makes the search more challenging. We name the space, which combines FPNetV2-F macro-architecture and FP-NAS micro-architecture, as \textit{FBNetV2-F++} space, which is $10^3\times$ larger than FBNetV2-F space.

\begin{table}[t]
\vspace{-0.6cm}
\centering
\resizebox{0.99\columnwidth}{!}{

  \begin{tabular}{c|c|c|c}
    \multirow{2}{*}{Micro-arch}  & \multicolumn{3}{c}{Search Variables}  \\
    & Kernel & Nonlinearity & No. of Splits \\
    \specialrule{.15em}{.1em}{.1em}
    
    FBNetV2 & \multirow{2}{*}{$\{0, 3, 5\}$}   &  \multirow{2}{*}{ $\{$\textit{relu}, \textit{swish}$\}$ }   & Squeeze-Excite  $\{0, 1\}$ \\
    FP-NAS & & & Split-Attention $\{0, 1, 2, 4\}$\\

  \end{tabular}
}
\caption{\textbf{FBNetV2 and FP-NAS micro architectures}. Kernel size and nonlinearity type are always searched. 
The difference is, for FBNetV2, it only searches whether Squeeze-Excite (SE) block is used. For FP-NAS, we search the number of splits in Split-Attention (SA) block. Choice 0 means SA block is not used, choice 1 means SE block is used, while choices 2 or 4 means SA block with 2 or 4 splits is used. For kernel size, choice 0 means we use a skip connection to bypass this layer to allow variable model depth.}
\label{tab:micro-structure}
\vspace{-0.2cm}
\end{table}

\begin{table}[t!]
\vspace{-0.2cm}
\centering
\resizebox{0.90\columnwidth}{!}{

  \begin{tabular}{c|cccc}
    Search space & Input size  & $\#$ Blocks & $\#$ Architectures & Median FLOPS (G)  \\
    \specialrule{.15em}{.1em}{.1em}
    FP-NAS-L0 &$224$ & 27 & $2\times 10^{32}$ & 0.39\\
    FP-NAS-L1 & $240$ & 32 & $1\times 10^{36}$ & 0.9 \\
    FP-NAS-L2 & $256$ &  32 & $1\times 10^{36}$ & 1.1 \\

  \end{tabular}
}
\caption{\textbf{FP-NAS spaces}. We define 3 FP-NAS spaces to search large models of different sizes.}
\label{tab:FPNAS spaces}
\vspace{-0.2cm}
\end{table}
\noindent{\textbf{FP-NAS spaces.}} The largest model in FBNetV2-F++ space only use $122$M FLOPS when input size is 128. To demonstrate the efficiency of our search method, we expand the FBNetV2-F macro-architecture in the following aspects. We increase the searchable channels in the block to make it wider. We also increase the repeating times of the block in the group to make it deeper. Last, we increase the input image size to classify the images at higher resolution for better accuracy. More details of the FP-NAS macro-architectures can be seen in the supplement. By combining the expanded macro-architectures and FP-NAS micro-architecture, we obtain three giant FP-NAS spaces L0-L2 (Table~\ref{tab:FPNAS spaces}), which contain models of different size for us to search. We also use FP-NAS-L to denote the searched models from these spaces.

\begin{figure*}[t!]
\vspace{-0.5cm}
   \centering
   \subfloat[]{\includegraphics[width=.22\textwidth]{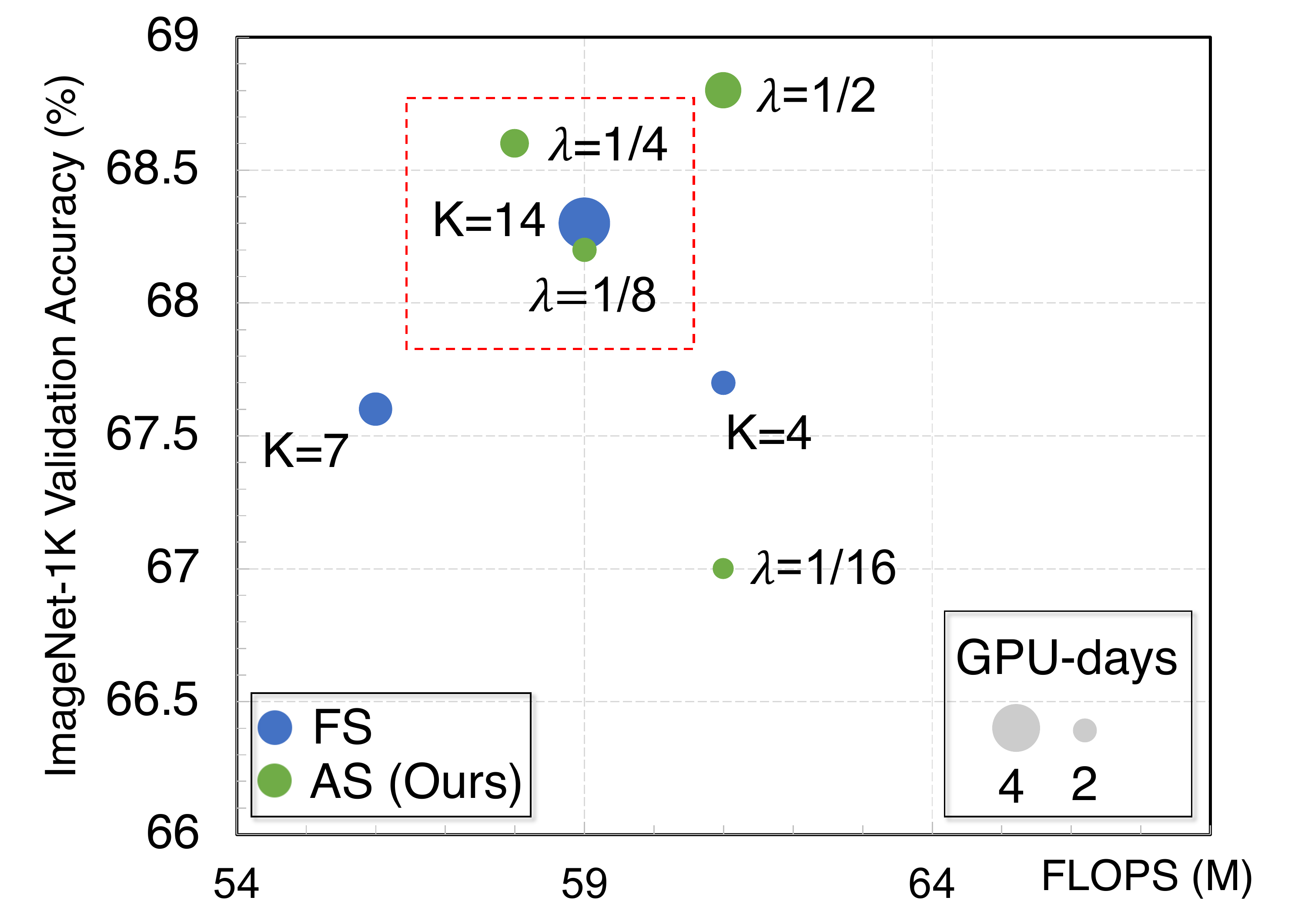} \label{fig:FS_AS_comp_acc_flops}}
   \subfloat[]{\includegraphics[width=.23\textwidth]{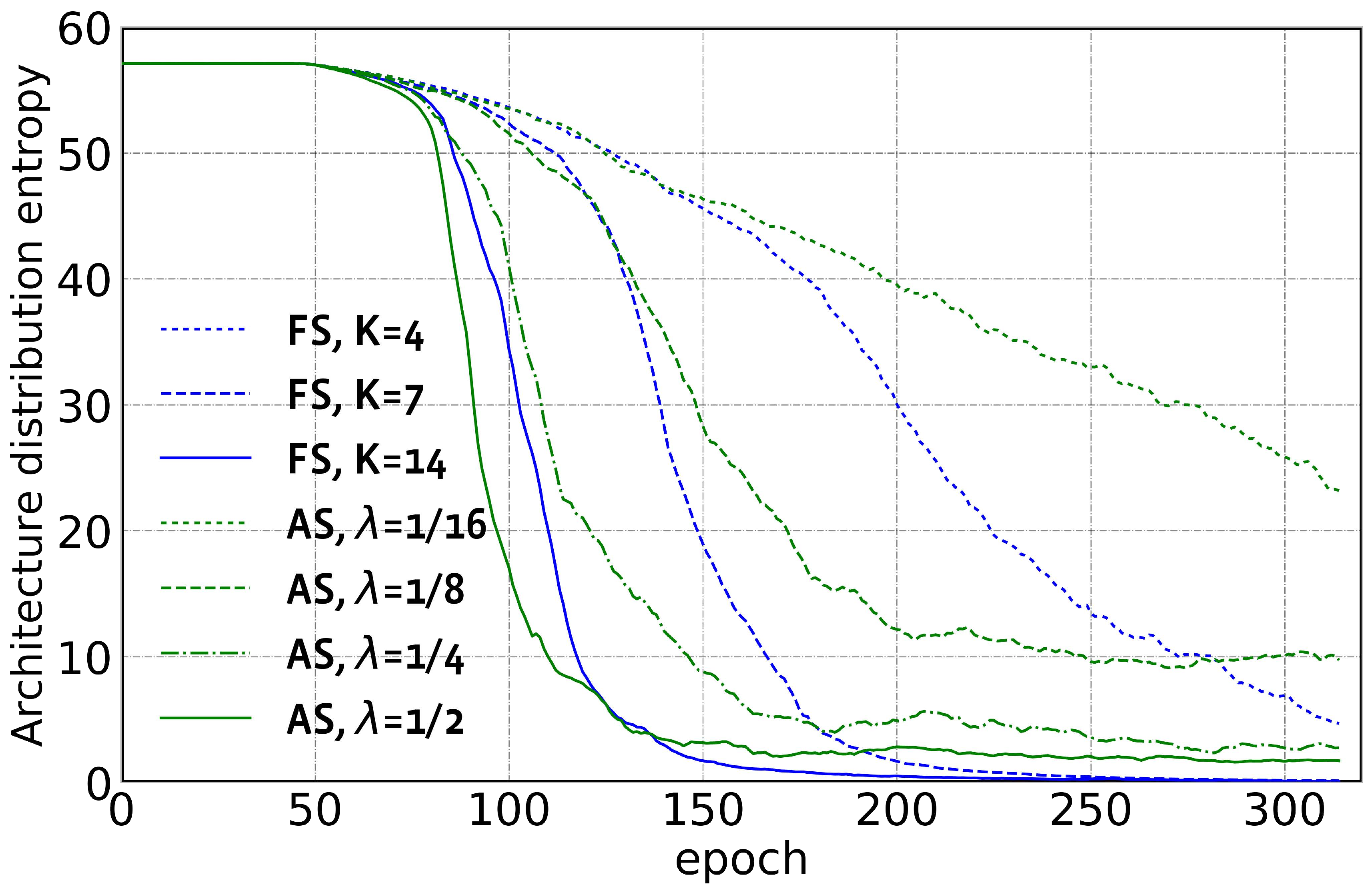}\label{fig:FS_AS_comp_entropy}}\quad 
   \subfloat[]{\includegraphics[width=.23\textwidth]{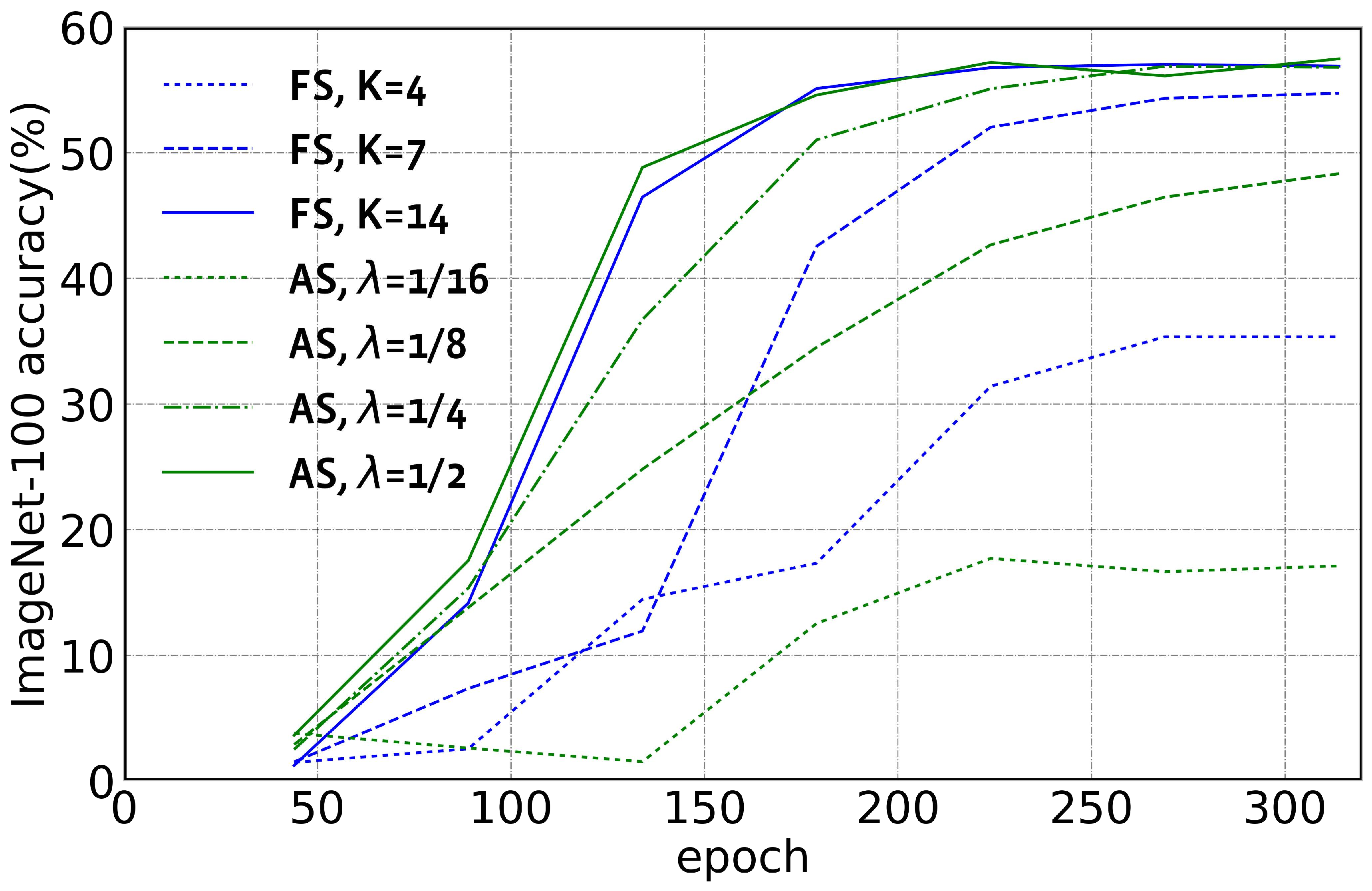}\label{fig:FS_AS_comp_val_acc}}\quad 
   \subfloat[]{\includegraphics[width=.25\textwidth]{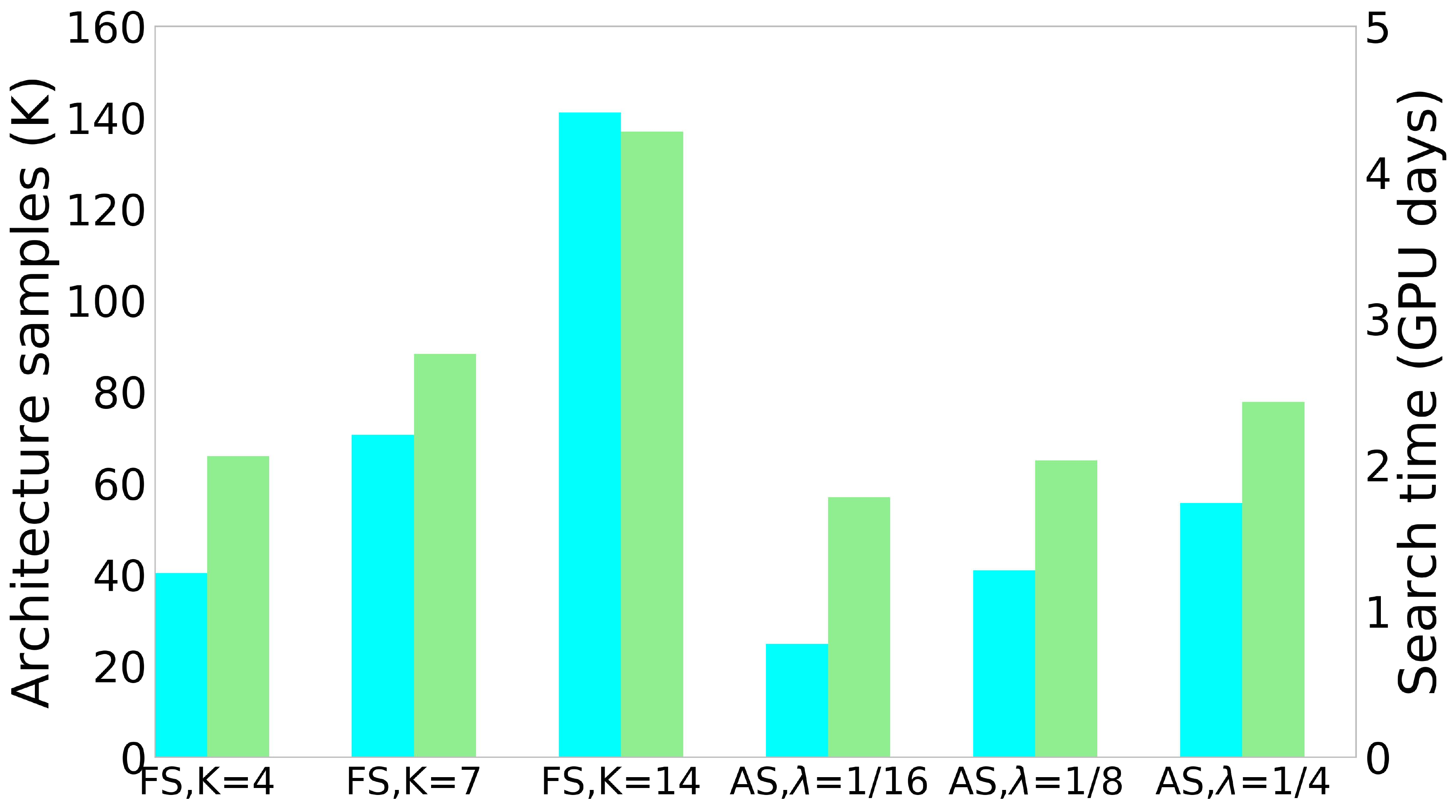} \label{fig:FS_AS_comp_search_time}}\quad 

   \caption{\textbf{Comparing sampling methods}. \textit{FS} and \textit{AS} denote fixed- and adaptive sampling. \textbf{(a)}: Accuracy and FLOPS of the final architecture on the ImageNet-1K validation set. The marker size encodes the search cost in GPU-days. We highlight the region where architectures with comparable accuracy-to-complexity trade-off are searched by FS and AS using significantly different GPU days.  \textbf{(b)}: Distribution entropy decreases after 45 warm-up epochs.  More samples $K$ or larger $\lambda$ lead to lower final entropy, but AS can reduce entropy as much as FS with fewer total samples.
   \textbf{(c)}: Accuracy of the most probable architectures on the ImageNet-100 validation set. Model weights are directly borrowed from the super-net. AS achieves high accuracy with fewer total architecture samples. \textbf{(d)}: Comparing the total sampled architectures and time cost of the search.}
   \label{fig:on_num_samples}
\vspace{-0.1cm}
\end{figure*}

\section{Experiments}
\label{sec:experiments}

\subsection{Implementation Details}
\vspace{-0.1cm}

We implement FP-NAS in PyTorch, and search architectures on 8 Nvidia V100 GPUs. We prepare the search dataset by randomly choosing 100 classes from ImageNet, namely \textit{ImageNet-100}. We use half of the original training set as training data to update model weights $\omega$, and the other half as the validation data to update model importance scores $\{m_k\}$ as well as architecture parameters $\alpha$. Our default hyper-parameters are as follows. To optimize $\alpha$, we use Adam~\cite{2014adam} with constant learning rate $0.016$. To optimize $\omega$, we use SGD with initial learning rate $0.8$ and follow a cosine schedule. Batch size is 256 images per GPU. We search architectures for 315 epochs, where the beginning 45 epochs are used to warm-up the supernet and only model weights $\omega$ are updated. The coefficient of cost penalty $\beta$ is set to $0.3$, and the coefficient $\lambda$ in adaptive sampling is set to $\frac{1}{4}$.

We search small models in FBNetV2-F, FBNetV2-F-Fine and FBNetV2-F++ space. We also search large models in FP-NAS spaces. After search, the final model is evaluated on the full ImageNet dataset. We use 64 V100 GPUs to train it from scratch for 420 epochs, using RMSProp with momentum $0.9$.  The initial learning rate is 0.2, and decays by $0.9875$ per epoch. We adopt Auto-Augment~\cite{2017autoaugment}, label smoothing~\cite{2016label-smooth}, Exponential Model Averaging and stochastic depth~\cite{2016droppath} to improve the training as in prior work~\cite{2019FBNetV2, 2019efficientnet}. More detailed comparisons on the training recipe can be found in the supplement. Finally, we report the top-1 validation accuracy.

\subsection{The Effectiveness of Adaptive Sampling}
\label{sec:adaptive_sampling}

\subsubsection{How Many Samples Should We Draw?}
\label{sec:on_arch_sample_size}

The original PARSEC~\cite{2019pnas} uses fixed sampling (FS), and constantly draws $K$ architectures (e.g. 8 or 16). Below we conduct a study in the FPNetV2-F space to show the choice of $K$ has a significant impact on the search. For FS, in Fig~\ref{fig:FS_AS_comp_acc_flops}, there is a strong correlation between $K$ and the final architecture quality in terms of Accuracy-To-Complexity (ATC) trade-off on ImageNet-1K validation set. In Fig~\ref{fig:FS_AS_comp_entropy}, a larger $K$ samples more architectures, and the distribution entropy is reduced more substantially, which means learning of the architecture distribution is more effective. In Fig~\ref{fig:FS_AS_comp_val_acc}, we show the ImageNet-100 validation accuracy of the most probable architecture at the end of each search epoch. The joint optimization of architecture parameters and model weights is more effective with a larger $K$. More samples help to better estimate the gradients, and leads to a faster learning of the distribution, which in turn samples promising architectures more often, and focuses more on updating model weights associated with them. 
In Fig~\ref{fig:FS_AS_comp_search_time}, the total sampled architectures and the search time by FS with $K=14$ is $3.5\times$ and $2.1\times$ more compared with those by FS with $K=4$, indicting the computational cost of the search with FS increases almost linearly in $K$. 

We also experiment adaptive sampling (AS) with different $\lambda \in \{\frac{1}{16}, \frac{1}{8}, \frac{1}{4}, \frac{1}{2} \}$. AS adjusts the sample size on the fly. For example, AS with $\lambda=\frac{1}{4}$ draws 14 samples in the beginning, on par with FS with $K=14$. However, as distribution entropy decreases, it will reduce the samples to save computation, and only draw a single sample at the end of search. In Fig~\ref{fig:FS_AS_comp_acc_flops}, AS with
$\lambda=\frac{1}{8}$ can search an architecture with ATC trade-off similar to that of the one from FS with $K=14$ using much fewer GPU-days. A larger choice of $\lambda=\frac{1}{4}$ for AS further improves the the ATC trade-off. A further larger choice of $\lambda=\frac{1}{2}$ for AS does not improve the ATC trade-off, but will increases the search time. Therefore, we use $\lambda=\frac{1}{4}$ in the following experiments. In Fig~\ref{fig:FS_AS_comp_entropy}, AS with larger $\lambda$ samples more architectures, and reduces distribution entropy faster. Both $\lambda=\frac{1}{4}$ and $\frac{1}{2}$ can reduce the entropy to a low level at the end. In Fig~\ref{fig:FS_AS_comp_val_acc}, AS with both $\lambda=\frac{1}{4}$ and $\frac{1}{2}$ can achieve the high final validation accuracy on ImageNet-100 comparable to that of FS with $K=14$. In Fig~\ref{fig:FS_AS_comp_search_time}, \textit{AS with $\lambda=\frac{1}{4}$ samples $60\%$ fewer architectures, and searches $1.8\times$ faster compared with FS with $K=14$}.

\subsubsection{FP-NAS Adapts to Search Space Size}

\begin{table}[t!]
\small
\begin{center}
\resizebox{0.92\columnwidth}{!}{

\begin{tabular}{cc|ccc}
  \multicolumn{2}{c}{Search Space}  & Sampling &  Model & Top-1 \\
  Name & $\#$ Architectures &  Method &  FLOPS (M) & Acc ($\%$)\\
  
  \specialrule{.15em}{.1em}{.1em}      
   \multirow{2}{*}{FBNetV2-F} & \multirow{2}{*}{ $6\times10^{25}$ } & FS, $K=14$ & 56 & 68.3\\
   & & AS, $\lambda=0.25$ & 58 & 68.6 \\ 
   \hline
   \multirow{2}{*}{FBNetV2-F-Fine} & \multirow{2}{*}{ $1\times10^{45}$ } & FS, $K=14$  &50 & 66.3\\
   & & AS, $\lambda=0.25$ & 51 & 67.2 \\    

\specialrule{.1em}{.05em}{.05em}

\end{tabular}

}
\end{center}
\vspace{-0.2cm}
\caption{\textbf{Comparison of models searched in small and large space using two different sampling methods.}}
\label{tab:adaptive_sampling}
\end{table}

In larger search spaces, there are much more architecture choices which requires to draw more samples to explore the space and learn the distribution. For FS, using a constant sample size $K$ will only discover sub-optimal models in larger search space. In contrast, AS with a constant value of $\lambda$ will adjust the sample size based on the distribution entropy, and does not require manual hyper-parameter tuning. To see this, for FS with $K=14$ and AS with $\lambda=0.25$, we compare the searched models from FBNetV2-F and FBNetV2-F-Fine space, where the latter is $10^{8}\times$ larger. The results are shown in Table~\ref{tab:adaptive_sampling}. In small FBNetV2-F space, the models discovered by two sampling methods have comparable ATC trade-off. However, in larger FBNetV2-F-Fine space, without changing the hyper-parameter of each sampling method, \textit{the search with AS discovers a significantly better model with $0.9\%$ higher accuracy}.

\subsection{Fast Coarse-to-Fine Search}
\label{sec:coarse_fine_search}

\begin{figure}[t!]
   \centering
   \subfloat[]{\includegraphics[width=.47\columnwidth]{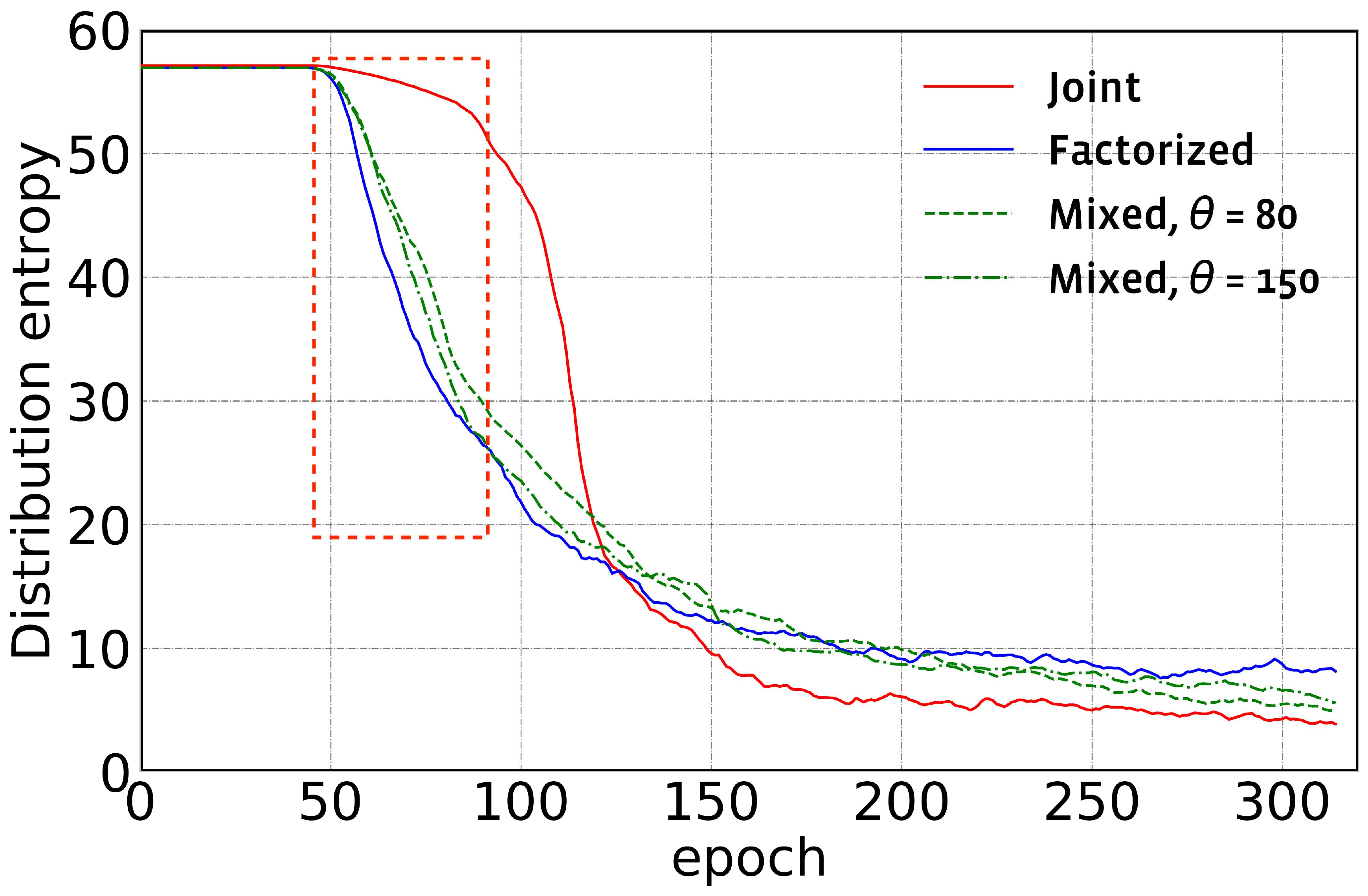}\label{fig:dist_schedule_entropy}}\quad 
   \subfloat[]{\includegraphics[width=.43\columnwidth]{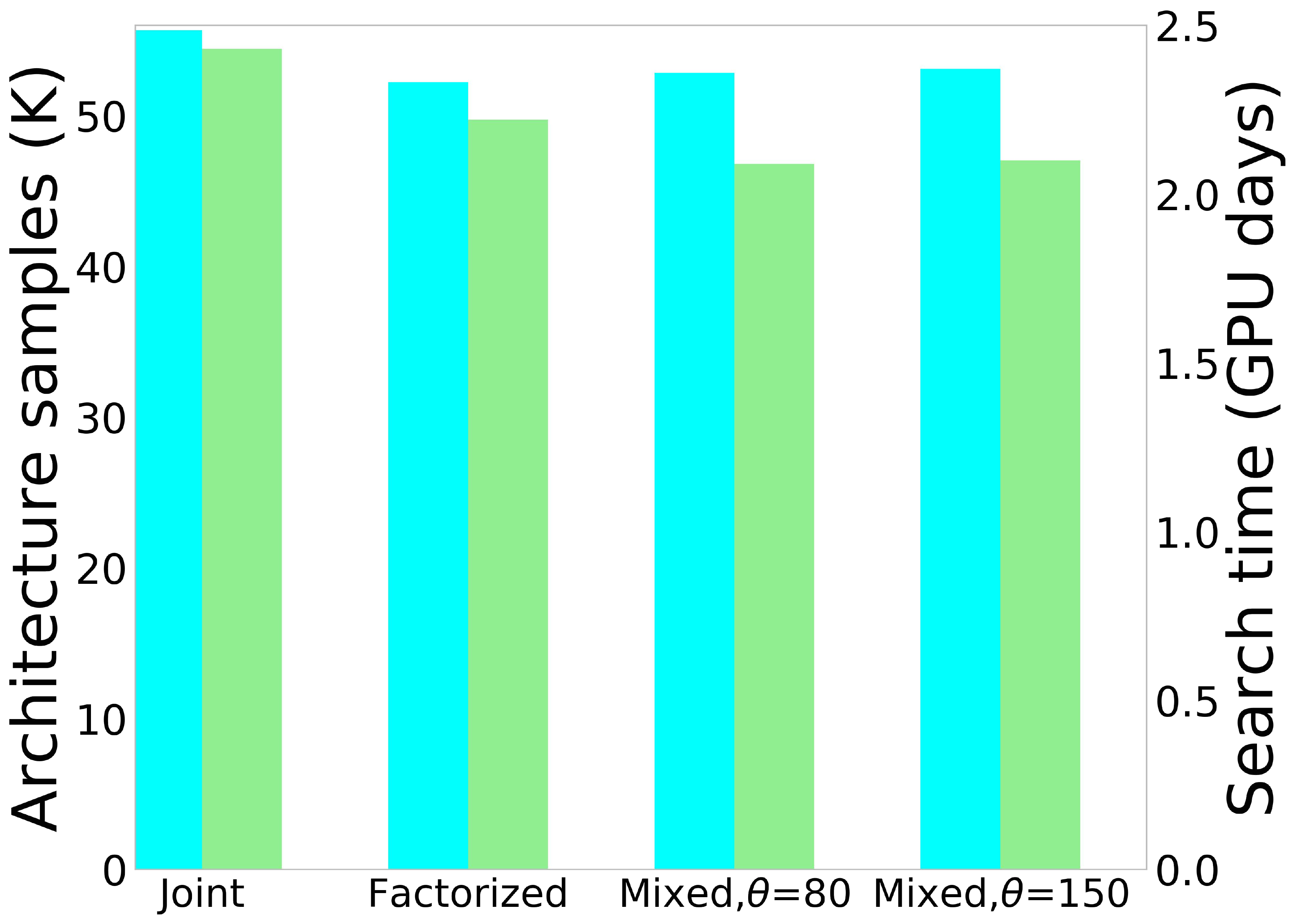}\label{fig:dist_schedule_search_time}}
  \vspace{-0.1cm}
   \caption{\textbf{Comparing joint-, factorized- and our proposed mixed distributions.} \textbf{(a):} The architecture distribution entropy during the search. The beginning part of the search, where the entropy of the factorized distribution is reduced much faster than that of the joint distribution, is highlighted by the dashed box. \textbf{(b):} Total architecture samples and overall search time, which are highly correlated, for different choices of architecture distribution scheduling. }
   \label{fig:coarse_to_fine}
\vspace{-0.1cm}
\end{figure}

In Fig~\ref{fig:coarse_to_fine}, we first compare the search with joint distribution (JD) only and factorized distribution (FD) only. The search with FD can reduce the entropy much faster than the search with the JD. The entropy at epoch 80 is quite different (30.6 Vs. 54.4). But in the later stage of the search, where the entropy is lower but not yet converged, the search with FD reduces the entropy slower than that with the JD, which means it struggles to distinguish architectures among a smaller set of candidates at a finer granularity. 

In our proposed coarse-to-fine search, we use a schedule of mixed distributions (MD), by starting the search with FD, and later convert it to the JD at search epoch $\theta$.  In Fig~\ref{fig:dist_schedule_entropy}, we also show results of the search with MD using two different $\theta \in \{80, 150\}$, and also compare with the baseline search using either JD only or FD only.  The search with MD can reduce the entropy almost as fast as that with FD at the beginning of the search. After FD is converted into JD for more fine-grained search, it can further reduce the entropy nearly as low as that in the search with JD only. Since the number of sampled architectures is proportional to the entropy of the distribution in our adaptive sampling, the search, which has faster reduction in the distribution entropy, samples fewer architectures, executes fewer forward/backward passes for sampled architectures, and eventually runs faster. 

In Fig~\ref{fig:dist_schedule_search_time} and Table~\ref{tab:coarse_to_fine}, we show the coarse-to-fine search reduces architecture samples by $9\%$, runs $1.2\times$ faster than the search using JD only, and does not hurt ATC trade-off. \textit{Compared with the original PARSEC, the FP-NAS search with both adaptive sampling and the schedule of mixed distribution reduces samples by $64\%$ and runs $2.1\times$ faster}.

\begin{table}[t!]
\vspace{-0.1cm}
\small
\begin{center}
\resizebox{0.9\columnwidth}{!}{

\begin{tabular}{cccc}
  Distribution & Architecture & Search Cost  & Model \\
  Schedule & $\#$ Samples (K) & (GPU days)  & Top-1 Acc ($\%$)\\
  
  \specialrule{.15em}{.1em}{.1em}      
  \textbf{Joint only} & 59.9 & 2.4 & 68.6\\
  \textbf{Factorized only} & 54.4 (-5.5) & 2.2 (-0.2)  & 67.6  (-1.0) \\ 
  \textbf{Mixed, $\theta$=80} & 54.3 (-5.6) & 2.1 (-0.3)  & 68.6\\
  \textbf{Mixed, $\theta$=150} & 51.8 (-8.1) & 2.1 (-0.3) & 68.2 (-0.4)\\

\specialrule{.1em}{.05em}{.05em}

\end{tabular}

}
\end{center}
\vspace{-0.2cm}

\caption{\textbf{Comparing the schedule of  architecture distribution and the accuracy of the searched models.}. Adaptive sampling with $\lambda=\frac{1}{4}$ is always used when different distribution schedules are compared. We report the top-1 accuracy on ImageNet-1K validation set. The search space FBNetV2-F is used. All models in the table use a comparable amount of FLOPS between 58-60M FLOPS.}
\label{tab:coarse_to_fine}
\vspace{-0.2cm}
\end{table}

\subsection{Comparisons with FBNetV2}

We search 4 small models in FBNetV2-F space using target FLOPS 60M, 90M, 130MF, and 250MF, respectively, and name them as FP-NAS-S models. We compare with 4 FBNetV2 models (i.e. from S1 to S4), which are searched in the same space. Results are shown in Table~\ref{tab:comp_fbnetv2}. Our method not only searches $1.9\times$ to $3.6\times$ faster, but also discovers models with better ATC trade-off. This demonstrates the superior search effectiveness and efficiency of our method. Furthermore, we stress that FBNetV2 method \textit{can not} be used to search large models due to its excessively large memory footprint which is required to cache all choices of layer operation. In contrast, our FP-NAS method uses much smaller memory footprint and can be used to directly search large models, as we will show later in Section~\ref{sec:search_large_model}.

\begin{table}[t!]
\vspace{-0.6cm}
\small
\begin{center}
\resizebox{0.82\columnwidth}{!}{

\begin{tabular}{c|c|cc}
  Input size & Model & Search cost (GPU days) &  Top-1 Acc ($\%$)\\
  
  \specialrule{.15em}{.1em}{.1em}      

  \multirow{2}{*}{128} & FBNetV2-F1 & 8.3 & 68.3 \\
  & \textbf{FP-NAS-S1 (ours)}  & \textbf{2.4 (-5.9)} & \textbf{68.6 (+0.3)}\\
  \midrule
  \multirow{2}{*}{160} & FBNetV2-F2 & 8.3  & 71.1 \\
  & \textbf{FP-NAS-S2 (ours)}  & \textbf{2.8 (-5.5) }  & \textbf{71.3 (+0.2)}\\
  \midrule
  \multirow{2}{*}{192} & FBNetV2-F3 & 8.3  & 73.2 \\
  & \textbf{FP-NAS-S3 (ours)}  & \textbf{3.3 (-5) }  & \textbf{73.4 (+0.2)}\\
  \midrule
  \multirow{2}{*}{256} & FBNetV2-F4 & 8.3  & 75.9 \\
  & \textbf{FP-NAS-S4 (ours)} & \textbf{4.3 (-4)} & \textbf{76.2 (+0.3)} \\

\specialrule{.1em}{.05em}{.05em}

\end{tabular}

}
\end{center}
\vspace{-0.3cm}
\caption{\textbf{Comparisons with FBNetV2~\cite{2019FBNetV2}}. Given the same input size, the FBNetV2 models and FP-NAS models here use a similar amount of compute with difference less than 4M FLOPS.}
\label{tab:comp_fbnetv2}
\vspace{-0.2cm}
\end{table}

\subsection{Searchable Split-Attention Module}
We search in FBNetV2-F++ space, which includes searchable Split-Attention (SA) module in MBConv block, and denote the searched models as \textit{FP-NAS-S++} models. In Table~\ref{tab:search_sa}, we compare them to FP-NAS-S models, which are searched in FBNetV2-F space. We also prepare two variants of FP-NAS-S models, by uniformly replacing the searched SE module with SA module using 2 or 4 splits. The searched FP-NAS-S++ models use a varying number of splits in different MBConv blocks (see model details in the supplement), and achieve significantly better ATC trade-off than FP-NAS-S models and their variants. This highlights the importance of searching the places of inserting SA modules and the number of splits for individual SA modules.

\begin{table}[t!]
\small
\begin{center}
\resizebox{0.9\columnwidth}{!}{

\begin{tabular}{c|c|ccc}
  Input & \multirow{2}{*}{Model} & Search Cost  & FLOPS & Top-1  \\
  Size &  &  (GPU days) & (M) &  Acc ($\%$)\\
  
  \specialrule{.15em}{.1em}{.1em}      

  \multirow{4}{*}{128} & FP-NAS-S1 & 2.4 & 58 & 68.6 \\
  & FP-NAS-S1, SA, uniform 2 splits & 2.4 & 65 & 69.3\\
  & FP-NAS-S1, SA, uniform 4 splits & 2.4 & 79 & 69.5\\
  & FP-NAS-S1++  & 4.4 & 66 & 70.0\\
  \midrule
  \multirow{4}{*}{160} & FP-NAS-S2 & 2.8 & 88 & 71.3 \\
  & FP-NAS-S2, SA, uniform 2 splits & 2.8 & 99 & 71.8 \\
  & FP-NAS-S2, SA, uniform 4 splits & 2.8 & 120 & 72.4 \\  
  & FP-NAS-S2++  & 4.8 & 98 & 72.2 \\
  \midrule
  \multirow{4}{*}{192} & FP-NAS-S3 & 3.3 & 131 & 73.4 \\
  & FP-NAS-S3, SA, uniform 2 splits & 3.3 & 141  & 73.6 \\
  & FP-NAS-S3, SA, uniform 4 splits & 3.3 & 170 & 74.0 \\    
  & FP-NAS-S3++  & 5.8 & 147 & 74.2 \\
  \midrule
  \multirow{4}{*}{256} & FP-NAS-S4 & 4.3 & 240 & 76.2 \\
  & FP-NAS-S4, SA, uniform 2 splits & 4.3 & 262 & 76.4 \\
  & FP-NAS-S4, SA, uniform 4 splits & 4.3 & 307 & 76.6 \\     
  & FP-NAS-S4++ & 7.3 & 268 & 76.6  \\

\specialrule{.1em}{.05em}{.05em}

\end{tabular}

}
\end{center}
\caption{\textbf{Comparison of models searched in FBNetV2-F and FBNetV2-F++ space.}}
\label{tab:search_sa}
\vspace{-0.5cm}
\end{table}

\subsection{Searching For Large Models}
\label{sec:search_large_model}

To demonstrate the scalability of our method, we also search large models in FP-NAS spaces. Specifically, we search 3 models in FP-NAS spaces with different target GFLOPS $\{0.4, 0.7, 1.0\}$, and the searched models are denoted as \textit{FP-NAS-L} models. 
The results are shown in Table~\ref{tab:sota} and Fig~\ref{fig:flops_acc_plot}, where we compare them with EfficientNet models of similar FLOPS. While both EfficientNet-B0 and FP-NAS-L0 models are searched from scratch, \textit{our search runs at least $\FpnasEffnetSpeedup\times$ faster and FP-NAS-L0 achieve $0.7\%$ higher top-1 accuracy on ImageNet}. Different from EfficientNet-B1/B2, which are obtained by scaling up the EfficientNet-B0 models, our FP-NAS-L1/L2 models are searched from scratch, and improve the accuracy by $0.9\%$ and $0.4\%$, while reducing the search time by over an order of magnitude.

\begin{table}[t!]
\vspace{-0.6cm}
\small
\begin{center}
\resizebox{0.98\columnwidth}{!}{

\begin{tabular}{c|ccccc}
  \multirow{2}{*}{Model} & Search Cost & FLOPS & Params & \multirow{2}{*}{Distill} & Top-1\\ 
   &  (GPU days) & (M) & (M) & & Acc ($\%$)\\
  
  \specialrule{.15em}{.1em}{.1em}      

  MobileNetV3-Small~\cite{2019mobilenetv3} & $>$3,790\ddag & 66 & 2.9 & $\times$ & 67.4 \\
  FBNetV2-F1~\cite{2019FBNetV2}  & 8.3 & 56 & 6.1 & $\times$ & 68.3  \\
  \textbf{FP-NAS-S1++ (ours)} & 6.7 & 66 & 5.9 & $\times$ & \textbf{70.0} \\
  \midrule
  MobileNeXt-1.0~\cite{mobilenext} & - & 300 & 3.4 & $\times$ & 74.0 \\
  MobileNetV3-Large~\cite{2019mobilenetv3} & $>$3,790\ddag & 219 & 5.4 & $\times$ & 75.2 \\
  MnasNet-A1~\cite{2019Mnasnet} & 3,790\dag & 312 & 3.9 & $\times$ & 75.2 \\
  FBNetV2-F4~\cite{2019FBNetV2}  & 8.3 & 242 & 7.1 & $\times$ & 75.8  \\
  \multirow{2}{*}{BigNAS-S~\cite{bignas}}  & \multirow{2}{*}{-} & \multirow{2}{*}{242} & \multirow{2}{*}{4.5} & $\times$ & 75.3 \\
  &  &  &  & $\checkmark$ & 76.5 \\    
  \textbf{FP-NAS-S4++ (ours)} & 7.6 & 268 & 6.4 & $\times$ & \textbf{76.6} \\
  \midrule
  ProxylessNAS~\cite{2018proxylessnas} &  8.3 & 465 & - & $\times$ & 75.1\\
  MobileNeXt-1.1~\cite{mobilenext} & - & 420 & 4.28 & $\times$ & 76.7 \\
  EfficientNet-B0~\cite{2019efficientnet} & $>$3,790\ddag & 390 & 5.3 & $\times$ & 77.3  \\
  FBNetV2-L1~\cite{2019FBNetV2} & 25 & 325 & - & $\times$ & 77.2  \\
  AtomNas~\cite{2019atomnas} & 34 & 363 & 5.9 & $\times$ & 77.6 \\ 
  \multirow{2}{*}{BigNAS-M~\cite{bignas} } & \multirow{2}{*}{-}  & \multirow{2}{*}{418} & \multirow{2}{*}{5.5} & $\times$ & 77.4 \\
  & &  &  & $\checkmark$ & 78.9 \\ 
  \textbf{FP-NAS-L0 (ours)} & \Lzerosearchtime & 399 & 11.3 & $\times$ & \textbf{78.0}\\
  \midrule
  EfficientNet-B1~\cite{2019efficientnet} & $>$3,790\ddag & 734 & 7.8 & $\times$ & 79.1 \\ 
  \multirow{2}{*}{BigNAS-L~\cite{bignas}}  & \multirow{2}{*}{-} & \multirow{2}{*}{586} & \multirow{2}{*}{6.4} & $\times$ & 78.2 \\
  & &  &  & $\checkmark$ & 79.5 \\   
 \multirow{2}{*}{\textbf{FP-NAS-L1 (ours)}}     & \multirow{2}{*}{58.6} & \multirow{2}{*}{728} & \multirow{2}{*}{15.8} & $\times$ & \textbf{80.0}\\
  &  &  &  & $\checkmark$ & \textbf{80.9}\\
  \midrule
  EfficientNet-B2~\cite{2019efficientnet} & $>$3,790\ddag & 1,050 & 9.2 & $\times$ & 80.3 \\ 
  \multirow{2}{*}{BigNAS-XL~\cite{bignas}}  & \multirow{2}{*}{-} & \multirow{2}{*}{1,040} & \multirow{2}{*}{9.5} & $\times$ & 79.3 \\
  & &  &  & $\checkmark$ & 80.9 \\ 
  \multirow{2}{*}{\textbf{FP-NAS-L2 (ours)}} & \multirow{2}{*}{69.1} & \multirow{2}{*}{1,045} & \multirow{2}{*}{20.7} & $\times$ & \textbf{80.7} \\
  & & & & $\checkmark$ & \textbf{81.6} \\

\specialrule{.1em}{.05em}{.05em}

\end{tabular}

}
\end{center}
\caption{\textbf{Comparisons with other methods}.
\dag:search cost is based on the experiments in MnasNet. \ddag: MobileNetV3 and EfficientNet combines search methods from MnasNet and NetAdapt~\cite{2018Netadapt}. Thus, MnasNet search time  can be viewed as a lower bound of their search time.}
\label{tab:sota}
\vspace{-0.6cm}
\end{table}

\subsection{Comparisons with Other Methods}
\vspace{-0.1cm}
 FP-NAS can natively search both small and large models. We use simple distillation~\cite{hinton2015distilling}, where the large EfficientNet-B4 model is used as the teacher model, to further improve our L1 and L2 models. In Table~\ref{tab:sota} and Figure~\ref{fig:flops_acc_plot}, we compare FP-NAS models with others. Our models has shown significantly better ATC trade-off than others. We also compare to BigNAS~\cite{bignas} models with and without using inplace distillation~\cite{slimmablenetwork} in Table~\ref{tab:sota}. For small model, FP-NAS-S4++ without distillation already works as well as the BigNAS-S model with advanced inplace distillation. \textit{For large model, FP-NAS-L2 with vanilla distillation can outperform BigNAS-XL with inplace distillation by $0.7\%$ using less FLOPS}.

\section{Conclusions}
\vspace{-0.3cm}

We presented a fast version of the probabilistic NAS. We demonstrate its superior performance by directly searching architectures, including both small and large ones, in large spaces, and validate their high performance on ImageNet.

{\small
\bibliographystyle{ieee_fullname}
\bibliography{egbib}

\begin{thebibliography}{10}\itemsep=-1pt

\bibitem{bender2018understanding}
Gabriel Bender, Pieter-Jan Kindermans, Barret Zoph, Vijay Vasudevan, and Quoc
  Le.
\newblock Understanding and simplifying one-shot architecture search.
\newblock In {\em International Conference on Machine Learning}, pages
  550--559, 2018.

\bibitem{2018proxylessnas}
Han Cai, Ligeng Zhu, and Song Han.
\newblock Proxylessnas: Direct neural architecture search on target task and
  hardware.
\newblock {\em arXiv preprint arXiv:1812.00332}, 2018.

\bibitem{2000monte-carlo}
Bradley~P Carlin and Thomas~A Louis.
\newblock Empirical bayes: Past, present and future.
\newblock {\em Journal of the American Statistical Association},
  95(452):1286--1289, 2000.

\bibitem{2019pnas}
Francesco~Paolo Casale, Jonathan Gordon, and Nicolo Fusi.
\newblock Probabilistic neural architecture search.
\newblock {\em arXiv preprint arXiv:1902.05116}, 2019.

\bibitem{2019DATA-NAS}
Jianlong Chang, Yiwen Guo, GAOFENG MENG, SHIMING XIANG, Chunhong Pan, et~al.
\newblock Data: Differentiable architecture approximation.
\newblock In {\em Advances in Neural Information Processing Systems}, pages
  876--886, 2019.

\bibitem{2017autoaugment}
Ekin~D Cubuk, Barret Zoph, Dandelion Mane, Vijay Vasudevan, and Quoc~V Le.
\newblock Autoaugment: Learning augmentation strategies from data.
\newblock In {\em Proceedings of the IEEE conference on computer vision and
  pattern recognition}, pages 113--123, 2019.

\bibitem{2019chamnet}
Xiaoliang Dai, Peizhao Zhang, Bichen Wu, Hongxu Yin, Fei Sun, Yanghan Wang,
  Marat Dukhan, Yunqing Hu, Yiming Wu, Yangqing Jia, et~al.
\newblock Chamnet: Towards efficient network design through platform-aware
  model adaptation.
\newblock In {\em Proceedings of the IEEE Conference on computer vision and
  pattern recognition}, pages 11398--11407, 2019.

\bibitem{2019oneshot4hr}
Xuanyi Dong and Yi Yang.
\newblock Searching for a robust neural architecture in four gpu hours.
\newblock In {\em Proceedings of the IEEE Conference on computer vision and
  pattern recognition}, pages 1761--1770, 2019.

\bibitem{du2020spinenet}
Xianzhi Du, Tsung-Yi Lin, Pengchong Jin, Golnaz Ghiasi, Mingxing Tan, Yin Cui,
  Quoc~V Le, and Xiaodan Song.
\newblock Spinenet: Learning scale-permuted backbone for recognition and
  localization.
\newblock In {\em Proceedings of the IEEE/CVF Conference on Computer Vision and
  Pattern Recognition}, pages 11592--11601, 2020.

\bibitem{2016resnet}
Kaiming He, Xiangyu Zhang, Shaoqing Ren, and Jian Sun.
\newblock Deep residual learning for image recognition.
\newblock In {\em Proceedings of the IEEE conference on computer vision and
  pattern recognition}, pages 770--778, 2016.

\bibitem{hinton2015distilling}
Geoffrey Hinton, Oriol Vinyals, and Jeff Dean.
\newblock Distilling the knowledge in a neural network.
\newblock {\em arXiv preprint arXiv:1503.02531}, 2015.

\bibitem{2019mobilenetv3}
Andrew Howard, Mark Sandler, Grace Chu, Liang-Chieh Chen, Bo Chen, Mingxing
  Tan, Weijun Wang, Yukun Zhu, Ruoming Pang, Vijay Vasudevan, et~al.
\newblock Searching for mobilenetv3.
\newblock In {\em Proceedings of the IEEE International Conference on Computer
  Vision}, pages 1314--1324, 2019.

\bibitem{2018senet}
Jie Hu, Li Shen, and Gang Sun.
\newblock Squeeze-and-excitation networks.
\newblock In {\em Proceedings of the IEEE conference on computer vision and
  pattern recognition}, pages 7132--7141, 2018.

\bibitem{2017densenet}
Gao Huang, Zhuang Liu, Laurens Van Der~Maaten, and Kilian~Q Weinberger.
\newblock Densely connected convolutional networks.
\newblock In {\em Proceedings of the IEEE conference on computer vision and
  pattern recognition}, pages 4700--4708, 2017.

\bibitem{2016droppath}
Gao Huang, Yu Sun, Zhuang Liu, Daniel Sedra, and Kilian~Q Weinberger.
\newblock Deep networks with stochastic depth.
\newblock In {\em European conference on computer vision}, pages 646--661.
  Springer, 2016.

\bibitem{2016gumbel-trick}
Eric Jang, Shixiang Gu, and Ben Poole.
\newblock Categorical reparameterization with gumbel-softmax.
\newblock {\em arXiv preprint arXiv:1611.01144}, 2016.

\bibitem{2014adam}
Diederik~P Kingma and Jimmy Ba.
\newblock Adam: A method for stochastic optimization.
\newblock {\em arXiv preprint arXiv:1412.6980}, 2014.

\bibitem{2012alexnet}
Alex Krizhevsky, Ilya Sutskever, and Geoffrey~E Hinton.
\newblock Imagenet classification with deep convolutional neural networks.
\newblock In {\em Advances in neural information processing systems}, pages
  1097--1105, 2012.

\bibitem{2017hierarchical-nas}
Hanxiao Liu, Karen Simonyan, Oriol Vinyals, Chrisantha Fernando, and Koray
  Kavukcuoglu.
\newblock Hierarchical representations for efficient architecture search.
\newblock {\em arXiv preprint arXiv:1711.00436}, 2017.

\bibitem{2018darts}
Hanxiao Liu, Karen Simonyan, and Yiming Yang.
\newblock Darts: Differentiable architecture search.
\newblock {\em arXiv preprint arXiv:1806.09055}, 2018.

\bibitem{2018shufflenetv2}
Ningning Ma, Xiangyu Zhang, Hai-Tao Zheng, and Jian Sun.
\newblock Shufflenet v2: Practical guidelines for efficient cnn architecture
  design.
\newblock In {\em Proceedings of the European conference on computer vision
  (ECCV)}, pages 116--131, 2018.

\bibitem{2019atomnas}
Jieru Mei, Yingwei Li, Xiaochen Lian, Xiaojie Jin, Linjie Yang, Alan Yuille,
  and Jianchao Yang.
\newblock Atomnas: Fine-grained end-to-end neural architecture search.
\newblock {\em arXiv preprint arXiv:1912.09640}, 2019.

\bibitem{2019xnas}
Niv Nayman, Asaf Noy, Tal Ridnik, Itamar Friedman, Rong Jin, and Lihi Zelnik.
\newblock Xnas: Neural architecture search with expert advice.
\newblock In {\em Advances in Neural Information Processing Systems}, pages
  1977--1987, 2019.

\bibitem{2018param-share-nas}
Hieu Pham, Melody~Y Guan, Barret Zoph, Quoc~V Le, and Jeff Dean.
\newblock Efficient neural architecture search via parameter sharing.
\newblock {\em arXiv preprint arXiv:1802.03268}, 2018.

\bibitem{2019AmoebaNet}
Esteban Real, Alok Aggarwal, Yanping Huang, and Quoc~V Le.
\newblock Regularized evolution for image classifier architecture search.
\newblock In {\em Proceedings of the aaai conference on artificial
  intelligence}, volume~33, pages 4780--4789, 2019.

\bibitem{2017large-scale-evolution}
Esteban Real, Sherry Moore, Andrew Selle, Saurabh Saxena, Yutaka~Leon Suematsu,
  Jie Tan, Quoc Le, and Alex Kurakin.
\newblock Large-scale evolution of image classifiers.
\newblock {\em arXiv preprint arXiv:1703.01041}, 2017.

\bibitem{2018mobilenetv2}
Mark Sandler, Andrew Howard, Menglong Zhu, Andrey Zhmoginov, and Liang-Chieh
  Chen.
\newblock Mobilenetv2: Inverted residuals and linear bottlenecks.
\newblock In {\em Proceedings of the IEEE conference on computer vision and
  pattern recognition}, pages 4510--4520, 2018.

\bibitem{2017inceptionv4}
Christian Szegedy, Sergey Ioffe, Vincent Vanhoucke, and Alex Alemi.
\newblock Inception-v4, inception-resnet and the impact of residual connections
  on learning.
\newblock {\em arXiv preprint arXiv:1602.07261}, 2016.

\bibitem{2015inceptionv1}
Christian Szegedy, Wei Liu, Yangqing Jia, Pierre Sermanet, Scott Reed, Dragomir
  Anguelov, Dumitru Erhan, Vincent Vanhoucke, and Andrew Rabinovich.
\newblock Going deeper with convolutions.
\newblock In {\em Proceedings of the IEEE conference on computer vision and
  pattern recognition}, pages 1--9, 2015.

\bibitem{2016label-smooth}
Christian Szegedy, Vincent Vanhoucke, Sergey Ioffe, Jon Shlens, and Zbigniew
  Wojna.
\newblock Rethinking the inception architecture for computer vision.
\newblock In {\em Proceedings of the IEEE conference on computer vision and
  pattern recognition}, pages 2818--2826, 2016.

\bibitem{2019Mnasnet}
Mingxing Tan, Bo Chen, Ruoming Pang, Vijay Vasudevan, Mark Sandler, Andrew
  Howard, and Quoc~V Le.
\newblock Mnasnet: Platform-aware neural architecture search for mobile.
\newblock In {\em Proceedings of the IEEE Conference on Computer Vision and
  Pattern Recognition}, pages 2820--2828, 2019.

\bibitem{2019efficientnet}
Mingxing Tan and Quoc~V Le.
\newblock Efficientnet: Rethinking model scaling for convolutional neural
  networks.
\newblock {\em arXiv preprint arXiv:1905.11946}, 2019.

\bibitem{2019FBNetV2}
Alvin Wan, Xiaoliang Dai, Peizhao Zhang, Zijian He, Yuandong Tian, Saining Xie,
  Bichen Wu, Matthew Yu, Tao Xu, Kan Chen, et~al.
\newblock Fbnetv2: Differentiable neural architecture search for spatial and
  channel dimensions.
\newblock In {\em Proceedings of the IEEE/CVF Conference on Computer Vision and
  Pattern Recognition}, pages 12965--12974, 2020.

\bibitem{2018FBNet}
Bichen Wu, Xiaoliang Dai, Peizhao Zhang, Yanghan Wang, Fei Sun, Yiming Wu,
  Yuandong Tian, Peter Vajda, Yangqing Jia, and Kurt Keutzer.
\newblock Fbnet: Hardware-aware efficient convnet design via differentiable
  neural architecture search.
\newblock In {\em Proceedings of the IEEE Conference on Computer Vision and
  Pattern Recognition}, pages 10734--10742, 2019.

\bibitem{2017resnext}
Saining Xie, Ross Girshick, Piotr Doll{\'a}r, Zhuowen Tu, and Kaiming He.
\newblock Aggregated residual transformations for deep neural networks.
\newblock In {\em Proceedings of the IEEE conference on computer vision and
  pattern recognition}, pages 1492--1500, 2017.

\bibitem{2018Netadapt}
Tien-Ju Yang, Andrew Howard, Bo Chen, Xiao Zhang, Alec Go, Mark Sandler,
  Vivienne Sze, and Hartwig Adam.
\newblock Netadapt: Platform-aware neural network adaptation for mobile
  applications.
\newblock In {\em Proceedings of the European Conference on Computer Vision
  (ECCV)}, pages 285--300, 2018.

\bibitem{slimmablenetwork}
Jiahui Yu and Thomas~S Huang.
\newblock Universally slimmable networks and improved training techniques.
\newblock In {\em Proceedings of the IEEE International Conference on Computer
  Vision}, pages 1803--1811, 2019.

\bibitem{bignas}
Jiahui Yu, Pengchong Jin, Hanxiao Liu, Gabriel Bender, Pieter-Jan Kindermans,
  Mingxing Tan, Thomas Huang, Xiaodan Song, Ruoming Pang, and Quoc Le.
\newblock Bignas: Scaling up neural architecture search with big single-stage
  models.
\newblock {\em arXiv preprint arXiv:2003.11142}, 2020.

\bibitem{2020resnest}
Hang Zhang, Chongruo Wu, Zhongyue Zhang, Yi Zhu, Zhi Zhang, Haibin Lin, Yue
  Sun, Tong He, Jonas Mueller, R Manmatha, et~al.
\newblock Resnest: Split-attention networks.
\newblock {\em arXiv preprint arXiv:2004.08955}, 2020.

\bibitem{2018shufflenet}
Xiangyu Zhang, Xinyu Zhou, Mengxiao Lin, and Jian Sun.
\newblock Shufflenet: An extremely efficient convolutional neural network for
  mobile devices.
\newblock In {\em Proceedings of the IEEE conference on computer vision and
  pattern recognition}, pages 6848--6856, 2018.

\bibitem{mobilenext}
Daquan Zhou, Qibin Hou, Yunpeng Chen, Jiashi Feng, and Shuicheng Yan.
\newblock Rethinking bottleneck structure for efficient mobile network design.
\newblock {\em ECCV, August}, 2020.

\bibitem{2016nas}
Barret Zoph and Quoc~V Le.
\newblock Neural architecture search with reinforcement learning.
\newblock {\em arXiv preprint arXiv:1611.01578}, 2016.

\bibitem{2019nasnet}
Barret Zoph, Vijay Vasudevan, Jonathon Shlens, and Quoc~V Le.
\newblock Learning transferable architectures for scalable image recognition.
\newblock In {\em Proceedings of the IEEE conference on computer vision and
  pattern recognition}, pages 8697--8710, 2019.

\end{thebibliography}
}

\clearpage 
\appendix
\renewcommand{\thefigure}{\thesection.\arabic{figure}}    
\renewcommand{\thetable}{\thesection.\arabic{table}} 

\end{document}